\documentclass[11pt]{article}
\usepackage{geometry}
\usepackage{coling2020}
\usepackage{times}
\usepackage{url}
\usepackage{latexsym}
\usepackage{microtype}
\usepackage{hhline}
\usepackage{booktabs}

\colingfinalcopy 

\title{UPB at SemEval-2020 Task 11: Propaganda Detection with Domain-Specific Trained BERT}

\author{Andrei Paraschiv,  Dumitru-Clementin Cercel, Mihai Dascalu \\
   University Politehnica of Bucharest, Faculty of Automatic Control and Computers \\
  {\tt andrei.paraschiv74@stud.acs.pub.ro} \\
  \tt\{dumitru.cercel, mihai.dascalu\}@upb.ro}
  
\date{}

\begin{document}
\maketitle
\begin{abstract}
Manipulative and misleading news have become a commodity for some online news outlets and these news have gained a significant impact on the global mindset of people. Propaganda is a frequently employed manipulation method having as goal to influence readers by spreading ideas meant to distort or manipulate their opinions. This paper describes our participation in the SemEval-2020, Task 11: Detection of Propaganda Techniques in News Articles competition. Our approach considers specializing a pre-trained BERT model on propagandistic and hyperpartisan news articles, enabling it to create more adequate representations for the two subtasks, namely propaganda Span Identification (SI) and propaganda Technique Classification (TC). Our proposed system achieved a F1-score of 46.060\% in subtask SI, ranking 5th in the leaderboard from 36 teams and a micro-averaged F1 score of 54.302\% for subtask TC, ranking 19th from 32 teams.
\end{abstract}

\section{Introduction}
\label{intro}

\blfootnote{This work is licensed under a Creative Commons Attribution 4.0 International Licence. Licence details: http://creativecommons.org/licenses/by/4.0/.
}

Propagandistic texts can be perceived as deliberate acts of spreading ideas with the self-serving purpose of influencing, distorting, and manipulating the opinions of other individuals or groups \cite{EMNLP19DaSanMartino}. Propagandistic texts use several techniques to achieve these goals - for example, using emotionally charged statements to persuade the audience not on the merits of the argument, but based on the visceral reaction to it. Thus, early detection and the labeling of such texts is of utmost importance to lessen the negative impact on the society and diminish the spread of such news through online social media.

SemEval-2020 Task 11 proposes the detection of propagandistic text sequences in English news articles, including a span detection subtask (subtask SI), followed by a fine-grained propaganda classification into 14 classes (subtask TC) \cite{DaSanMartinoSemeval20task11}. The competition was divided into three phases. Participants had access to the training set with full golden labels and a dev set without the golden labels. During the development phase, participants could submit the results on the dev set for both the SI and TC subtasks, and receive an instant score. In contrast, the test set submissions received the scores only at the end of the competition.

Our solution builds upon BERT \cite{devlin2018bert}, a pre-trained Transformer-based \cite{vaswani2017attention} language model composed of several (12 or 24) multi-head attention blocks. The pre-trained model provided by Google was further pre-trained to create a domain-specific model using masked language modeling on a corpus consisting of news articles. This approach was used in narrow domains, like scientific papers \cite{beltagy2019scibert} or bio-medical data \cite{lee2020biobert}, with significant performance gain in domain-specific tasks. 

The rest of the paper is organized as follows. Section 2 presents work related to the propaganda detection problem. Section 3 provides details on the dataset, its processing, together with model implementation details. Afterwards, we discuss the achieved results in section 4, followed by conclusions in the last section. 

\section{Related Work}

Propaganda detection is performed in general on article-level information and is regarded as a subarea of fake news detection. Rashkin et al.~\shortcite{rashkin2017truth} show that the level of truthfulness across several domains (such as news articles, online statements, or opinions) exhibits common linguistic attributes; moreover, stylistic cues can contribute to identifying the truth of certain news articles. Thus, lexical features can be used for automated classification of news into different types, like fake news, propaganda, satire, or hoaxes. \newcite{volkova2017separating} achieved an accuracy of 95\% in classifying tweets into suspicious or verified. They also achieved a micro F1-score of 92\% and a macro F1-score of 72\%
for distinguishing between propaganda, satire, hoaxes, or click-bait by using additional cues from the text and also by encoding the social graph of interactions for the Twitter accounts.

In addition, the shared tasks of previous editions of SemEval tackled the issue of hyperpartisan news \cite{kiesel2019semeval} with remarkable results. Approaches ranged from traditional NLP models like Support Vector Machines, k-nearest neighbor, Naive Bayes, Logistic Regression \cite{gupta2019clark}, n-grams \cite{nguyen2019nlp} to Transformer-based approaches \cite{drissi2019harvey}. Recently, \newcite{EMNLP19DaSanMartino} proposed a novel span-level annotation, where specific portions of a news are tagged with the corresponding propaganda technique from a predefined list of labels. This allows not only a binary classification into propaganda or non-propagandistic articles, but also a fine-grained classification into 18 identified propaganda techniques. \newcite{yoosuf2019fine} used a BERT-based approach on the dataset proposed by \newcite{EMNLP19DaSanMartino} and achieved a F1-score of 38.98\%  on the identification of the propagandistic spans and 22.58\% F1-score on classifying them into the proposed classes. 

\section{Method}

\subsection{Corpus}

The dataset for the SemEval-2020 Task 11 contained 371 training articles and 75 articles in the test set. The training set contained 6,129 propaganda labeled spans, with three over-represented classes (i.e., Loaded Language - 2,123 samples,  Name Calling - 1,058 samples, Repetition - 621 samples). For both subtasks, we split the training data into 297 training articles and 74 validation articles. Moreover, we used a balanced mix of 50\% - 50\% by undersampling the negative class due the number of non-propaganda samples which was almost three times larger than the ones containing propaganda. A challenge on the TC subtask was the high unbalance between classes (e.g., Bandwagon, Reductio ad hitlerum has 72 samples whereas Loaded Language has 2123 samples). 

\subsection{Text Splitting}
We experimented with two different text splitting approaches for the SI subtask. First,  each article was split into paragraphs, and both the training and test batches were created as sequences of paragraphs. Since the rows in the training batches were shuffled, by keeping a whole paragraph, we tried to preserve more local context. Some paragraphs could exceed the maximum sequence length of 128 tokens used  by us, longer paragraphs were also divided into two or more sub-paragraphs by considering the nearest punctuation mark before the 128th token. Second, articles are split directly into sentences and batches contain shuffled sentences. 

Regarding the TC subtask , additional information about the exact location of the spans was provided. This led to a third text splitting approach, namely the extraction of the exact text span, its labeling, and usage as a training sample. We also experimented with both paragraph and sentence splitting for the classification subtask, using only paragraphs or sentences that contained a propagandistic span in training, but the scores were significantly below the exact span splitting.  

\subsection{Models}
Our intuition is that propagandistic news and biased news rely on language-specific constructs, word co-occurrences, and vocabulary that are not necessarily captured by the broader pre-training datasets. Similar to models trained on domain-specific corpora~\cite{beltagy2019scibert}, our focus was to expose models to “biased news”. Thus, we further pre-trained Google released  BERT-base-uncased model (12 heads, 768 dimensions) and created our model (BERTp) using masked language modeling \cite{devlin2018bert} on a dataset comprising of two corpora: a) 8.5M fake and suspicious news articles released by opensources.co\footnote{\url{https://github.com/several27/FakeNewsCorpus}} and b) 750k articles from the hyperpartisan news corpus \cite{kiesel2019semeval}. For this purpose, we pre-processed the previously mentioned corpora and trained the BERTp model for 2 million steps, with 10,000 warm-up steps and a learning rate of 1e-4. 

We added for the SI subtask a 768-dimensional dense layer on top of the last (12th) Transformer block, followed by a Conditional Random Fields (CRF) layer~\cite{lafferty2001conditional} with two labels, 0 for non-propaganda and 1 for propaganda. Since propaganda spans are mostly much longer than a few words, we use the CRF layer for joint label decoding. The rectified Adam optimizer ~\cite{liu2019variance} with a 1e-3 learning rate is used for training. We compare our neural network with both (1) BiLSTM-CRF model \cite{huang2015bidirectional} using GloVe 300-dimensional embeddings \cite{pennington2014glove}, as well as (2) the original BERT model without further pre-training.

We tackle the multi-label classification task by adding a dense layer funnel with sizes of 768, 256, and 14, followed by a final softmax layer to estimate the probability distribution over the classes.
Moreover, we create an ensemble with majority voting using the BERTp model with five different pre-training checkpoints. Since propagandistic spans did overlap, we avoided labeling two overlapping spans with the same class by choosing the most probable class predicted by the softmax function. A rectified Adam optimizer with a 1e-4 learning rate was used, whereas our baseline for the TC subtask consists of: (1) a BiLSTM model with a dense-softmax classifier and GloVe 300-dimensional embeddings, (2) the original BERT model without further pre-training, and (3) one single instance of BERTp.

\section{Results}
Tables 1 and 2 present the results of the SI subtask. We can observe that all BERT-based models surpassed BiLSTM-CRF with a large margin on the development set. Also, the split did not impact the BERT model. However, our improved BERTp model could gain better precision by splitting the articles into larger chunks, while maintaining the same recall as for the sentence level split. Our best model (i.e., BERTp-CRF, ps) achieved a precision of 58.612\% and a recall of 37.936\%, totaling a F1-score of 46.06\% and reaching the 5th place from 36 teams on the leaderboard. 
\begin{table}[ht]
\centering
\begin{tabular}{llll}
\hline
Method & Precision (\%) & Recall (\%) & F1-score (\%)\\
\hline
BiLSTM-CRF, ps & 39.53 & 23.65 & 29.59\\
BERT-CRF, ps & \textbf{54.62} & 28.69 & 37.62\\
BERT-CRF, ss & 35.03 & 40.89 & 37.73\\
BERTp-CRF, ps & 42.44 & \textbf{52.13} & \textbf{46.79}\\
BERTp-CRF, ss & 29.23 & 52.95 & 37.66\\
  \hline
\end{tabular}
\caption{Model scores for the subtask SI on the dev set for the used models and the two types of article splitting (\textit{ps} denotes paragraph split, \textit{ss} means sentence split). }
\label{table:scoreMappigTable}
\end{table}

\begin{table}[ht]
\centering
\begin{tabular}{llll}
\hline
Method &   Precision (\%) & Recall (\%) & F1-score (\%)\\
\hline
Best ranked submission (Team Hitachi) & 56.544 & 47.368 & 51.551 \\
BERTp-CRF, ps (Ours) & 58.612 & 37.936 & 46.060 \\
  \hline
\end{tabular}
\caption{Model scores for the subtask SI on the test set for our best model, with input texts split into paragraphs.}
\label{table:scoreMappigTableTestSet}
\end{table}

We experimented with several article splitting strategies for the subtask TC, on a sentence or paragraph basis, but the best performance was achieved by taking only the propagandistic span as training data. The additional context did not support a better classification (see Table \ref{table:scoreTaskTC}). Oversampling of less frequent classes was used to overcome the class imbalance problem. Also, during the BERT pre-training phase, we saved intermediary snapshots of the model after certain number of steps. Thus, we experimented with several majority voting ensembles formed from checkpoints saved during the pre-training phase and subsequently, each of them fine-tuned on the labeled data. The best improvement was achieved by using the checkpoints at 350k, 800k, 1.2M, 1.5M, and 2M steps. As depicted in Table \ref{table:scoreTaskTCDetail_1}, the ensemble model had better performance on both minority (such as "Bandwagon, Reductio ad hitlerum" and "Repetition") and majority classes (e.g., "Loaded Language" and "Name Calling, Labeling"). Our best model achieved a micro-average F1 score of 54.302\% on the test data, placing us in 19th place from 32 teams.

\begin{table}[ht]
\centering
\begin{tabular}{ll}
\hline
Method & $\mu$F1 (\%) \\
\hline
BiLSTM & 49.57\\
BiLSTM, oversampling & 46.56\\
BERT & 54.94\\
BERT, oversampling & 53.71\\
BERTp & 56.91\\
BERTp, oversampling & 52.39\\
BERTp, ensemble & \textbf{58.33}\\

\hline
\end{tabular}
\caption{Micro-averaged F1 ($\mu$F1) scores on the development set for subtask TC.}
\label{table:scoreTaskTC}
\end{table}

\begin{table}[ht]
\centering
\begin{tabular}{p{5.12cm}p{1.2cm}p{0.90cm}p{1.2cm}p{1.6cm}|p{1.7cm}p{1.45cm}}

          \hline
            & Spans Per Label & BERT (\%)  & BERTp (\%) & BERTp Ensemble (\%) & Our final Submission (\%)& Best 
   Team (\%)\\
\hline
Appeal to Authority  & 144 & 5.56 & 7.14 & \textbf{7.41} & 20.00 & 48.15\\
Appeal to fear-prejudice & 294  & \textbf{36.56} & 28.24 & 29.89 & 30.00 & 45.49\\
Bandwagon, Reductio ad hitlerum & 72 & 40.00 & 33.33 & \textbf{57.14} & 0.00 & 8.33 \\
Black-and-White Fallacy & 107 & \textbf{18.75} & 8.70 & 8.70 & 19.72 & 49.02 \\
Causal Oversimplification & 209 & \textbf{37.84} & 35.29 & 34.29 & 16.95 & 22.73 \\
Doubt & 493  & \textbf{48.92} & 45.96 & 46.15 & 52.55 & 56.23\\
Exaggeration, Minimisation & 466 & 44.14 & \textbf{50.00} & \textbf{50.00} & 30.62  & 33.59\\
Flag-Waving  & 229 & 68.03 & \textbf{71.61} & 70.73 & 55.86 & 69.43\\
Loaded Language  & 2,123 & 71.15 & 70.22 & \textbf{74.02} & 70.09 & 77.12\\
Name Calling, Labeling & 1,058 & 68.46 & 70.23 & \textbf{71.64} & 68.86 & 74.38\\
Repetition  & 621 & 17.39 & 8.99  & \textbf{24.16} & 20.00 & 54.55 \\
Slogans & 129 & 44.74 & 42.11 & \textbf{50.00} & 34.61 & 51.28\\
Thought-terminating Cliches & 76 & \textbf{19.05} & 0.00 & 0.00 & 22.86 & 39.22\\
Whataboutism, Straw Men & 108 & 4.88 & \textbf{6.45} & 6.06 & 4.88 & 25.00\\
\hline
\end{tabular}
\caption{Scores $\mu$F1 for each of the subtask TC labels on both the development and test sets.}
\label{table:scoreTaskTCDetail_1}
\end{table}

\section{Conclusions}
This paper introduces our BERT-based model for both propaganda span detection, as well as propaganda technique classification. Further pretraining of the BERT model using large domain-specific corpora improves propaganda detection and classification performance. Experimenting with different text splitting methods has shown that even though for the original BERT model these have no significant influence, our BERTp model benefits from the larger context window.  The imbalanced dataset proved to be the largest challenge for the classification task. An ensemble with multiple checkpoints in the BERT pretraining process obtained better classification results for underrepresented classes.

In our future work we will analyze the impact of adding meta-features from named entity recognition on the performance of the propaganda span detection. Also, experiments with distilled versions of BERT, or BERT models with a reduced number of attention heads, are envisioned.

\bibliographystyle{coling}
\bibliography{semeval2020}

\end{document}